\documentclass[11pt,a4paper]{article}
\usepackage[hyperref]{acl2019}
\usepackage{times}
\usepackage{latexsym}
\usepackage{url}
\usepackage{graphicx} 
\usepackage{adjustbox} 
\usepackage{latexsym} 
\usepackage{amsmath} 
\usepackage{amssymb} 
\usepackage{paralist}  

\aclfinalcopy 

\title{Technical notes: Syntax-aware Representation Learning With Pointer Networks}

\author{Matteo Grella \\
  Turin, Italy \\
  \texttt{matteogrella@gmail.com} \\}

\date{}

\begin{document}
\maketitle
\begin{abstract}
This is a work-in-progress report, which aims to share preliminary results of a novel sequence-to-sequence schema for dependency parsing that relies on a combination of a BiLSTM and two Pointer Networks (Vinyals et al., 2015), in which the final softmax function has been replaced with the logistic regression. The two pointer networks co-operate to develop a latent syntactic knowledge, by learning the lexical properties of ``selection'' and the lexical properties of ``selectability'', respectively. At the moment and without fine-tuning, the parser implementation gets a UAS of 93.14\% on the English Penn-treebank (Marcus et al., 1993) annotated with Stanford Dependencies: 2-3\% under the SOTA but yet attractive as a baseline of the approach.\par
\end{abstract}

\section{Introduction}
The syntactic analysis via dependency parsing is considered to be a fundamental step for language processing because of its key importance in mediating between linguistic expression and meaning.\par
Modern approaches to dependency parsing (\newcite{dyer2015transitionbased}, \newcite{ballesteros2016dynamic}, \newcite{kiperwasser2016simple}, to name a few) use deep neural models as auxiliary components to the traditional transition-based and graph-based parsing algorithms  (Kubler et al., 2009).\par
As soon as the deep learning models proved to be successful in capturing the relevant information for the syntactic analysis, there has been a considerable increase of the number of parsing architectures in which the neural component is predominant, relaxing the needs of algorithmic constraints.\par
Among the recent parsing designs that differ from the traditional approaches, there are \newcite{dozat2016deep} who achieve state-of-the-art accuracies using a biaffine attention in a simple graph-based dependency parser; \newcite{DBLP:journals/corr/abs-1804-08915} and Li et al. (2019) who use a sequence-to-sequence schema which does not rely on any transition sequence by directly predicting the relative position of the head for each word in the sentence; \newcite{lhr2018} who use a bidirectional recurrent autoencoder to reconstruct for each i-word the j-word corresponding to its head in the sentence; Ma et al. (2018) who combine pointer networks to build the dependency tree in a top-down (from root-to-leaf) depth-first fashion; \newcite{2019arXiv190210505S} who use a sequence labeling strategy that outputs for each word the ``relative PoS-based encoding'' to find its head in the sentence.\par
The more a neural parser is independent of a superstructure\footnote{For \textit{independent of a superstructure}, we mean no transition-based framework, and in case of graph-based parsing the arcs scores are obtained before the search procedure.}, the more it is reasonable to think that the underlying neural model has learnt a ``syntactic knowledge'' such as to perform the task of dependency parsing at hand.\par
As a by-product of an encoder-decoder parsing schema, it is possible to use the internal parser encoded representation to boost the perfomance of other high-level tasks that benefit from syntactic information. For instance, \newcite{DBLP:journals/corr/abs-1804-08915} proposed a scheduled multi-task learning framework to train an encoder-decoder machine translation system sharing the encoder with a seq2seq dependency parser, concluding that syntactic auxiliary tasks are helpful not solely for machine translation but potentially for other systems as well.\par
Following the recent trend, this paper introduces a parsing model that relies on a combination of a BiLSTM and two Pointer Networks \cite{2015arXiv150603134V} over the linear sequence of tokens capable to hanlde unrestricted non-projective sentences. \par

The model is trained on two complementary syntactic tasks, as detailed in the sections below, with the aim to create a robust syntactic representation at the encoding layer. \par
At the moment and without fine-tuning, the parser implementation gets a UAS of 93.14\% on the English Penn-treebank (Marcus et al., 1993) annotated with Stanford Dependencies: 2-3\% under the SOTA but yet attractive as a baseline of the approach.\par
Extensive parsing evaluations, as well as experiments on the contribution that the dense representation resulting from the encoding layer could give to other high-level tasks, are still in the planning phase.

\section{Our Approach}
\label{sec:approach}

\subsection{Linguistic Background}
\label{sec:linguisticbackground}

Long-standing theories and formalisms (\newcite{Tesniere:1959}, \newcite{Sgall:1986}, \newcite{Mel'cuk:1988}, \newcite{hudson90}) share the fundamental assumption that syntactic structure consists of word-to-word dependencies, i.e., lexical nodes linked by binary asymmetrical relations. \par
More formally, dependencies can be represented as a set of directed arcs of the form \textit{g} $\xrightarrow{\textit{l}}$ \textit{d}, where \textit{g} is the head/governor node, \textit{d} is the dependent node (\textit{g} $\neq$ \textit{d}) and \textit{l} is the label, resulting in a dependency structure called dependency tree. Hence the name Dependency Grammar (DG). \par
The DG is sometimes called Valency Grammar, a name conceived by the analogy between the chemical valency and the thematic-argumental structure: a description of the elements that can depend on the word under consideration (its necessary complements, named arguments, and its optional complements, called modifiers).\footnote{Because of this analogy, sometimes it is possible to call the words ``atoms''.} \par
This possibility of a word to combine with other words selecting them as its dependents is hereby called \textbf{\textit{lexical selection property}}. The possibility of a word, instead, of being dependent on another word is hereby called \textbf{\textit{lexical selectability property}}. \par
We can therefore say that a well-formed sentence is a set of words combined so that the \textbf{\textit{selectability}} and \textbf{\textit{selection}} properties of each word are satisfied.\par
For more details on dependency tree, dependency grammar and dependency parsing see Nivre (2003) and the references cited therein.\par

\subsection{Neural Building Blocks}
\label{sec:neuralbuildingblocks}

Here is a quick overview of the main neural modules used in our approach. \par
See \newcite{goldberg-methods} for an extensive introduction of the neural building blocks used for the natural language processing.

\subsubsection{BiLSTM}

The BiLSTM \cite{graves2008supervised,irsoy2014opinion} consists in a bidirectional LSTMs \cite{hochreiter1997long} capable to learn bidirectional long-term dependencies between time steps of time series or sequence data. \par
The BiLSTM is a well established neural model used to represent the sentence tokens in their surrounding context.\footnote{\newcite{kiperwasser2016simple} were the first who demonstrated the effectiveness of using a conceptually simple BiLSTM in the context of dependency parsing.}

\subsubsection{Pointer Network}
The Pointer Network \cite{2015arXiv150603134V} is a type of neural network that works with a variable number of inputs and uses the attention mechanism (Bahdanau et al., 2015; Luong et al., 2015) to learn and predict the conditional probability of an output sequence, with elements that are discrete tokens corresponding to positions in an input sequence. \par
In other words, the pointer networks use a softmax output whose dimension is dynamic and corresponds to each input sequence in a such way that the output space is constrained to be the observation of the input sequence (not the input space), maximizing the attention probability of the target input.

\subsection{The Idea}
\label{sec:idea}

Like \newcite{zhang2016dependency} and other recent approaches, we formalize the dependency parsing as the task of finding for each word in a sentence its most probable head without tree structure constraints. \par

The particularity of our approach is that we consider the probablilty of a word\texttt{\textsubscript{i}} to be the head of another word\texttt{\textsubscript{j}} as the average of the probability of the word\texttt{\textsubscript{j}} (dependent) to be selected by the word\texttt{\textsubscript{i}} (governor), and the probability of the word\texttt{\textsubscript{i}} to select the word\texttt{\textsubscript{j}} as its dependent.\par
In fact, we want the neural model to learn the syntactic properties of ``selectability'' and ``selection'' more explicitly than other models that take into account only one of these two syntactic aspects. \par

To do this, we intuitively cast the main dependency parsing task in two sub-tasks: \textbf{heads-pointing} for the \textit{selectability} property and \textbf{dependents-pointing} for the \textit{selection} property. In essence, the task of \textbf{heads-pointing} consists to find the most probable \textit{head} of a given word in the sentence; the task of \textbf{dependents-pointing} consists to find the most probable \textit{dependents} of a given word in the sentence. \par The gist of the idea is that these sub-tasks should develop different views of the same problem and thus increase the robustness of the learnt \textit{syntactic knowledge}.\par

To solve the ``problems of pointing'' we decided to experiment with the Pointer Networks (Ptr-Net). The use of these networks in dependency parsing is not new: in Section \ref{sec:related} we have highlighted the main difference among the other models and our approach.\par

In the Ptr-Net, the output of the attention mechanism is a \textit{softmax distribution}, which allows to point to the heads because for each word there is only one governor according to the dependency grammar.\footnote{Even if it would force the use of a virtual root for the top node}
On the other side, since the dependents of a word can be more than one, the softmax function cannot be used to point to them, and this sets limits to the use of the Ptr-Net. \par

To overcome these limits we introduce a variant of the original Ptr-Net, in which the final \textit{softmax function} is replaced with the \textit{logistic regression}, so that independent predictions can enable multiple pointing at the same time as well as allowing no pointing at all. We call our variant Ind-Ptr-Net (Independent Pointer Network).\footnote{To date, the author is surprised not to have found any other references that deal with this variant of the Pointer Network.}\par

As detailed in section \ref{sec:processes}, the Ind-Ptr-Net is the backbone of our approach.\par

\subsection{Training and Inference Processes}
\label{sec:processes}

Our model is composed of one BiLSTM and two Ind-Ptr-Nets.\par

The BiRNN works as an encoder, receiving in input the tokens of a sentence (already transformed in a dense representation) and generating the \textit{context vectors}, that represent them in the sentence context. The \textit{context vectors} are in turn given as input sequence to the two Ind-Ptr-Nets.

Subsequently, the decoding process feeds again each \textit{context vector} into these Ind-Ptr-Nets to perform the \textbf{heads-pointing} task and \textbf{dependents-pointing} task respectively.\par

\paragraph{Training.}

During the training phase,

\begin{compactitem}
    \item[$-$] the Ind-Ptr-Net for the \textbf{heads-pointing} is trained to activate the output corresponding to the position of the head of the word under consideration (i.e., set it to 1.0). In case this word is the top, it is trained not to activate any output (i.e., set them all to zero).
    \item[$-$] the Ind-Ptr-Net for the \textbf{dependents-pointing} is trained to activate the outputs corrisponding to the positions of the dependents of the word under consideration. In case this word has no dependent, it is trained not to activate any output.
\end{compactitem}\par

The gradients are propagated from the two Ind-Ptr-Nets all the way back, through the BiLSTM until the initial tokens embeddings (which are trained together with the model).

\paragraph{Inference.}\par

During the inference phase, the outputs of the two Ind-Ptr-Nets are merged by averaging the attention values before the \textit{sigmoid} activation.\par

To construct the dependency tree we select, for each token, the head with the highest score. The top token of the sentence is found before assigning the other heads, looking for the token which has among all other tokens the pointers to the heads with the lowest scores (ideally, with all the scores equal to zero).\footnote{To construct a labeled dependency tree, it is possible to add a simple feedforward network that computes a classification of the labels giving in input the \textit{context-vectors} of each dependent-governor pair. However, we prefer to run further experiments before including any labeling results in this report.} \par 

At test time, we ensure that the dependency tree given in output is well-formed by iteratively identifying and fixing cycles with simple heuristics, without any loss in accuracy.\footnote{For each cycle, the fix is done by removing the arc with the lowest score and assigning to its dependent the node that maximizes its latent head similarity without introducing new cycles.} \par

We empirically observed that during the decoding most outputs are already trees, without the need to fix cycles. It seems to confirm once again that the linear sequence of tokens itself is sufficient to recover the underlying dependency structure (\newcite{zhang2016dependency}).

\section{Related Approaches}
\label{sec:related}
The use of Pointer Networks \cite{2015arXiv150603134V} in dependency parsing has been previously experimented by \newcite{2016chorowski} and \newcite{2019:3305347} who use the Ptr-Nets to predict the \textit{heads}, and \newcite{DBLP:journals/corr/abs-1805-01087} who use the Ptr-Nets to predict the \textit{dependents}. \par
The main difference with these two first approaches that ``point to the heads'', can be found on how the \textit{root} is selected, meaning that in our model it is not required a \textit{virtual element}: the top word is recognized as an emerging syntactic property because of the absence of strong connections with other words considered as heads. \par
The main difference with the approach that ``point to the dependents'', is that in our model it is not required to define a deterministic decoding process to select a dependent in multiple time steps, but all the dependents of a word can be pointed to simulteneously.

\section{Experiments and Results}
\label{sec:results}

The parser is implemented in Kotlin, using the SimpleDNN deep learning library\footnote{\url{https://github.com/KotlinNLP/SimpleDNN}}. The code will be released at the GitHub author repository soon. \par

A performance evaluation has been carried out on the Penn Treebank (PTB) \cite{penntb} converted to Stanford Dependencies (Marneffe et al., 2006) following the standard train/dev/test splits and without considering punctuation markers. This dataset contains a few non-projective trees. \par

Our baseline is obtained following the unlabeled parsing approach described in section \ref{sec:processes}. 

A good initial tokens encoding is crucial to obtain high results in neural parsing, especially for richly inflected languages.\footnote{For example, adding subword information with character-based representation to the words embedding has been shown to be effective enough to compensate the lack of POS tags information (\newcite{dozat2017stanford}).}\par
However, rather then top parsing accuracy, in this study we focus more on the ability of the proposed model to learn a latent representation capable to capture the information needed for the syntactic analysis; so, for our baseline, we opted for a simple encoded representation of the input tokens. \par

We encode the input tokens concatenating the vectors obtained from two embedding maps. The first associates the words found in the training-set with randomly initialized vectors; the second contains pre-trained word embeddings.\footnote{The pre-trained word embeddings are the same used in \newcite{dyer2015transitionbased} and \newcite{kiperwasser2016ef}; the random values are generated using the Glorot initialization.} Both maps are fine-tuned during the training phase.\par 

During the training we replace the embedding vector of a word with an ``unknown vector'' with a probability that is inversely proportional to the frequency of the word in the tree-bank (tuned with an $\alpha$ coefficient). \par

We optimize the parameters with the Adam update method \cite{kingma2014adam} with default parameters ($\alpha$ = 0.001 $\beta$ = 0.9 $\beta$ = 0.999). We performed a very minimal tuning of the hyper-parameters; the values used for our baseline are reported in Table \ref{tbl:hyper}. \par

We evaluated five different configurations of the parser (\textit{p}).

\begin{compactitem}
    \item[$p1$] the parser is trained to perform both the \textit{heads-pointing} and the \textit{dependents-pointing}. The scores of the pointers to the heads are the average of the scores resulting from the two sub-tasks;
    \item[$p2$] the parser is trained to perform both the \textit{heads-pointing} and the \textit{dependents-pointing}. The scores of the pointers result from the 1st task only;
    \item[$p3$] the parser is trained to perform both the \textit{heads-pointing} and the \textit{dependents-pointing}. The scores of the pointers result from the 2nd task only;
    \item[$p4$] the parser is trained to perform the \textit{heads-pointing} only;
    \item[$p5$] the parser is trained to perform the \textit{dependents-pointing} only;
\end{compactitem}\par

\bigbreak

We trained the four instances of the parsers with different random seeds up to 10 epochs, and for each parser we selected the model from the epoch with the best accuracy on the development set. The average of the experimental results in Table \ref{tbl:baseline}.

\begin{table}[ht]
  \begin{center}
    \begin{scalebox}{0.8}{
      \begin{tabular}{ | c | c | }
        \textbf{Hyper-param} & \textbf{Value} \\
        \hline 
        Pre-trained word embedding dimension & 100 \\
        Word embedding dimension & 150 \\
        IndPtrNets hidden dimension & 100 \\
        IndPtrNets hidden activation & Tanh \\
        IndPtrNets attention transformation & Affine \\
        IndPtrNets output activation & Sigmoid \\
        BiLSTMs activations & Tanh \\
        BiLSTMs levels & 2 \\
        $\alpha$ (word dropout) & 0.25  \\
        \hline
      \end{tabular}
    }\end{scalebox}
    \caption{Hyper-parameters used for the baseline.}
    \label{tbl:hyper}
  \end{center}
\end{table}

\begin{table}[ht]
    \centering
  \begin{center}
    \begin{scalebox}{0.8}{
      \begin{tabular}{ | c | c | c |}
        \textbf{Parser} & \textbf{Method} & \textbf{UAS} \\
        \hline 
        p1 (this work) & heads+deps (avg scores) & 93.14 (+0.27)\\
        p2 & heads+deps (heads scores) & 92.87 (+0.11)\\
        p3 & heads+deps (deps scores) & 92.76 (+0.32)\\
        p4 & heads & 92.44 (+0.07)\\
        p5 & deps & 92.37 \\
        \hline
      \end{tabular}
    }\end{scalebox}
    \caption{Evaluation of different parser configurations.}
    \label{tbl:baseline}
  \end{center}
\end{table}

This section will be updated with further experiments soon.

\subparagraph{Observation of the results:}
With these first experimental results (Table \ref{tbl:baseline}), we can observe that the two sub-tasks taken individually (\textit{p4} and \textit{p5}) get almost the same performance.\par As soon as the sub-tasks are trained together we can appreciate an increase in performance, even when in the inference phase only one of the two tasks (\textit{p3} or \textit{p2}) is considered. \par
When, in addition to the joint training, the average of the results of the two sub-tasks is calculated, a further increase of correct arcs is obtained (\textit{p1}).

\section{Conclusion and Future Works}

The main objective of this study is to verify the hypotesis that an explicit learning process that consider both the lexical properties of ``selectability'' and ``selection'' can result in a more ``aware'' syntactic representation. For this purpose, we are investigating what kind of ``knowledge of language'' the proposed neural model is capturing, extending the tests to grammaticality judgments and visualizing which information the networks consider more important in a given moment \cite{karpathy2015visualizing}.\footnote{In our experiments we found that the RAN \cite{lee2017recurrent} is a valid alternative to the LSTM of the bidirectional recurrent network, when speed and highly interpretable outputs are important.} 

In this paper we have introduced a simple encoder-decoder approach for dependency parsing that handles unrestricted non-projective dependencies naturally. \par 

We introduced a variant of the Pointer Network, named Ind-Ptr-Net (Independent Pointer Netwoek), where the final softmax function is replaced with the logistic regression, so that independent predictions can enable multiple pointing at the same time as well as allowing no pointing at all.\footnote{We also plan to experiment the use of the \textit{tanh} function: as the tanh has a derivative of up to 1.0, we think that ``larger updates'' of the weights can result in better and faster learning process.} \par 

With the aim of understanding the potential and the limits of the proposed approach we intend to test more sophisticated initial tokens encodings, and to evaluate the parsing model on other tree-banks with a higher ratio of non-projective sentences. \par

\bibliography{acl2019}
\bibliographystyle{acl_natbib}

\end{document}